\documentclass[letterpaper, 10 pt, journal, twoside]{IEEEtran}
\usepackage{xcolor}
\usepackage{soul}
\usepackage{multirow}
\usepackage{graphicx}
\usepackage{booktabs}
\usepackage{dirtytalk}
\usepackage{bm}
\usepackage{hyperref}
\usepackage{moresize}
\hypersetup{
    colorlinks=true,
    linkcolor=red,
    filecolor=magenta,      
    urlcolor=blue,
    pdftitle={Overleaf Example},
    pdfpagemode=FullScreen,
    }
\usepackage{amsmath}
\usepackage{amssymb}
\usepackage{bbold}
\usepackage{mathtools}
\usepackage{subcaption}
\usepackage{caption}
\usepackage{perl_acronyms}
\let\labelindent\relax
\usepackage[inline]{enumitem}
\usepackage{siunitx}
\usepackage[
    style=ieee,
    doi=false,
    isbn=false,
    url=false,
    eprint=false,
    backend=bibtex,
    natbib=true
    ]{biblatex}
\IEEEoverridecommandlockouts                            

\title{Beyond NeRF Underwater: Learning Neural Reflectance Fields for True Color Correction of Marine Imagery}

\author{Tianyi Zhang$^{1}$, Matthew Johnson-Roberson$^{1}$

\thanks{Manuscript received: April 10 2023; Revised: July 6, 2023; Accepted: July 28, 2023.}
\thanks{This paper was recommended for publication by Editor Cesar Cadena Lerma
 upon evaluation of the Associate Editor and Reviewers’ comments.}
\thanks{$^{1}$T. Zhang and  M. Johnson-Roberson are with the Robotics Institute, School of Computer Science,
        Carnegie Mellon University,
        Pittsburgh, PA 15213, USA
        {\tt\small \{tianyiz4, mkj\}@andrew.cmu.edu}}%
\thanks{ This work was supported by the National Oceanic and Atmospheric Administration (NOAA) under Grant NA22OAR0110624.}
}

\begin{document}

\maketitle

\begin{abstract}
\urlstyle{rm}
\renewcommand{\thefootnote}{\roman{footnote}}
Underwater imagery often exhibits distorted coloration as a result of light-water interactions, which complicates the study of benthic environments in marine biology and geography. In this research, we propose an algorithm to restore the true color (albedo) in underwater imagery by jointly learning the effects of the medium and neural scene representations. Our approach models water effects as a combination of light attenuation with distance and backscattered light.
The proposed neural scene representation is based on a neural reflectance field model, which learns albedos, normals, and volume densities of the underwater environment.
We introduce a logistic regression model
to separate water from the scene and apply distinct light physics during training. Our method avoids the need to estimate complex backscatter effects in water by employing several approximations, enhancing sampling efficiency and numerical stability during training. The proposed technique integrates underwater light effects into a volume rendering framework with end-to-end differentiability. Experimental results on both synthetic and real-world data demonstrate that our method effectively restores true color from underwater imagery, outperforming existing approaches in terms of color consistency.
Our code and data are released at \url{https://github.com/tyz1030/neuralsea.git}
\end{abstract}


\section{INTRODUCTION}
\label{sec:introduction}
\par \IEEEPARstart{O}{ptical} imaging is being widely used in exploring the benthic world together with modern underwater robotic systems. The visual information presented in RGB format reveals rich details about underwater ecosystems and artifacts. For example, images collected by an underwater robot can be used to assess the health of coral reefs and segment live corals from dead samples~\cite{mandersonAssess}. However, the colors displayed in underwater images are consistently distorted due to wavelength-dependent attenuation and veiling effects resulting from light-water interactions. Such effects alter the visual appearance of images, as well as the performance of downstream tasks such as detection, classification, or segmentation~\cite{funniegan}. Restoring the color in underwater imagery is of great interest to communities working on marine ecology, biology, and geography, etc.

\par The formation of underwater color distortion has seen significant work, in which two kinds of light-water interaction are commonly studied: attenuation and scattering~\cite{jaffe1990, mobley1994light}.
Attenuation describes the process whereby water absorbs light at varying rates depending on the wavelength. Red light is absorbed most quickly, leading to loss of the red part of the visual spectrum in typical underwater images~\cite{absorption}. Underwater light scattering refers to the process by which light is dispersed in various directions as it interacts with water molecules, suspended particles, and other microscopic elements within the underwater environment~\cite{jaffe1990}.
While in graphics multiple-scattering is typically modeled, in water photons reflected to the camera without striking the scene, i.e. backscatter, have a major impact on image formation by creating a veiling effect.
Although our understanding of water optics has advanced, restoring color in underwater images is still challenging. While these effects are well-modeled, accurately estimating them from real data in uncontrolled environments remains an open problem.

\begin{figure}[t]%
\centering
    \includegraphics[width=0.94\linewidth]{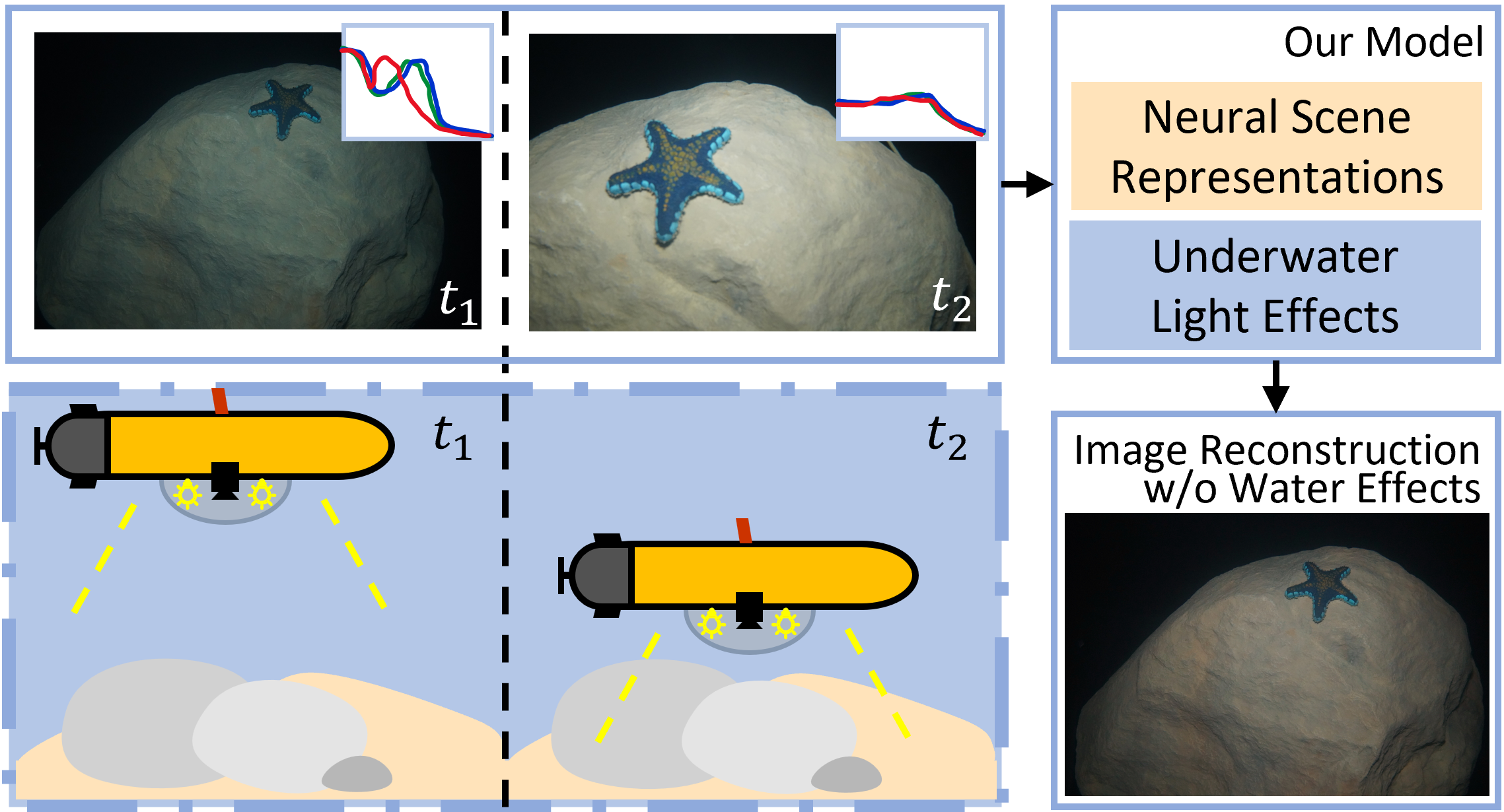}
    \caption{Observing an underwater scene from different altitudes results in variating color distribution over the RGB channels. Such observations encode the physics of light-water interactions. Our proposed model leverages this cue to restore the true color of underwater scenes by learning water effects together with neural scene representations.}
    \label{intro:concept}
\end{figure}

\par Early studies on marine optics developed underwater image formation models~\cite{jaffe1990, McGlamery} and measured absorption and scattering functions from different types of water samples~\cite{absorption, petzold1972volume}. With the above work, images can be synthesized with underwater effects~\cite{uwroborender}. However, this approach is insufficient for accurately correcting the color of real-world underwater images, as the measurements of a finite number of water optical properties cannot be reliably applied to novel field data.
Recent progresses on \ac{SfM} and deep learning have inspired the development of data-driven algorithms for underwater color correction.
\ac{SfM}-based method~\cite{ccbryson} estimates the true color (albedo) with multiple-view geometry constraints, but is only able to generate sparse results on feature points.
Deep-learning-based methods~\cite{Li2017WaterGANUG, funniegan, Sethuraman2022WaterNeRFNR} are able to correct the color with physical cues, but the result depends on prior color distributions or pre-training.
\par Combining insights from both types of methods, we developed a unified model that effectively restores the true color in underwater imagery (Fig.~\ref{intro:concept}). Our proposed model optimizes the attenuation and backscatter coefficients together with a neural reflectance field~\cite{neuralref} from a sequence of observations without any assumptions on prior color distributions. Based on the observation that water and scene are separable given volume density, we embed a logistic regression function in our neural scene representation which allows us to apply different light-transmitting physics to water and the scene, while maintaining end-to-end differentiability of our model. Our experiments demonstrate that our method is able to generate photorealistic results with restored true color in a dense format. Unlike previous studies that attempt to correct the color in the underwater images without taking lighting conditions into account, our work build physical model for underwater robots which have onboard light sources and cameras moving as a rigid body, thereby outperforming previous studies.

\section{RELATED WORK}
\label{sec:relatedwork}

\subsection{Underwater Image Formation Model}
According to {the} Jaffe-McGlamery model~\cite{jaffe1990, McGlamery}, the formation of underwater images can be decomposed into direct signals, forward-scattering, and backscatter. Direct signals refer to the light that is reflected from the underwater scene.
Backscatter refers to the phenomenon in which light enters a camera without being reflected directly from the scene. The trajectory of a photon after interacting with a particle in water is characterized by \acp{VSF}~\cite{petzold1972volume}. These empirical functions are dependent on both viewing and lighting directions.
Forward-scattering occurs when a photon deviates from its direct path before reaching the sensor, resulting in a blurred image.
This effect can be approximated by convolution operations~\cite{jaffe1990} or Gaussian blurring~\cite{uwroborender}.
\par In this work, to overcome the challenge of estimating \ac{VSF} with limited constraints from observations, {we propose several approximations applicable to underwater robots.}
Our scene representations do not model forward scattering, the error introduced by which is zero-mean and negligible~\cite{mildenhall2022rawnerf}.

\subsection{Neural Implicit Representations} Neural implicit representations
have been widely used in learning visual appearances and structures. NeRF~\cite{mildenhall2020nerf} is a kind of neural implicit representation that learns a 3D scene in the form of a neural field of volume density and radiance. The volume rendering equations in NeRF, which {are} based on \ac{RTE}, are not only good for inferring the 3D geometry of objects but also have the power to model the water effects such as absorption and scattering. Based on NeRF's framework, neural reflectance field~\cite{neuralref} and its {variants}~\cite{verbin2022refnerf, yang2022s3nerf} model the reflectance of the scene which enables the high-quality rendering under novel lighting conditions.
\par For underwater scenes illuminated by light sources attached to the robot, the appearance of the scene changes due to the robot's movement. To accommodate for these appearance changes resulting from varying illumination conditions, it is necessary to model reflectance properties of the scene. Therefore, we opt to use a neural reflectance field~\cite{neuralref} {backboned by iNGP}~\cite{mueller2022instant} as our foundational model.

\subsection{Underwater Color Correction} 
\par Early studies on underwater color correction make assumptions on underlying color distributions, e.g. histogram equalization~\cite{histeq}, grayworld~\cite{grayworld}, or dark-channel prior~\cite{darkchannel}. However, color balanced from the above assumptions lacks consistency when the same scene is observed from multiple views due to range-dependent water effects.
\par Bryson et al.~\cite{ccbryson} leverages the physical constraints from multiple-view geometry to estimate the true color of the scene. However, this method only estimates the true color of feature points and is unable to directly generate color-corrected images in a dense format.
\par Further progress in this field is made with deep learning approaches. WaterGAN~\cite{Li2017WaterGANUG} proposes to synthesize a dataset with ground truth depth and colors by training a GAN, then train a color correction network to restore the color. FUnIE-GAN~\cite{funniegan} emphasizes image quality for downstream tasks rather than adhering to physical constraints and, as such, can achieve real-time performance.
GAN-based methods, such as those mentioned above, require pre-training on a dataset. These methods can exhibit biases if the underlying color distribution differs from that of the training set.
In contrast, our approach does not require any pre-training on datasets. Rather, it restores color by creating scene representations using a series of observations from multiple perspectives.
\par WaterNeRF~\cite{Sethuraman2022WaterNeRFNR} utilizes mip-NeRF~\cite{barron2022mipnerf360} to model the underwater scene. Based on depth estimation from mip-NeRF, WaterNeRF learns the absorption and backscatter coefficients by optimizing the Sinkhorn loss between rendered image and histogram equalized image. Our approach diverges from WaterNeRF in that we model the scene as a reflectance field, which accounts for changes in illuminance, as opposed to a radiance field. Furthermore, we do not make any assumptions regarding the underlying color distributions.
\par Lastly, all the approaches mentioned above~\cite{Li2017WaterGANUG, Sethuraman2022WaterNeRFNR, ccbryson} use the model proposed in~\cite{scatprimemodel} to account for backscatter, which assumes natural and ambient light to be the major illumination source of scattering. In other words, their formulations are based on the assumption that the intensity of scattering is spatially constant, which does not hold for underwater robots equipped with light sources, taking light fall-off into consideration. In our work, we depart from~\cite{scatprimemodel}'s model and propose several approximations for underwater robots.

\section{METHODOLOGY}
\label{sec:methodology}
\subsection{Neural Scene Representation}
\begin{figure*}[t]%
\centering
    \begin{subfigure}{.80\textwidth}
    \includegraphics[width=\linewidth]{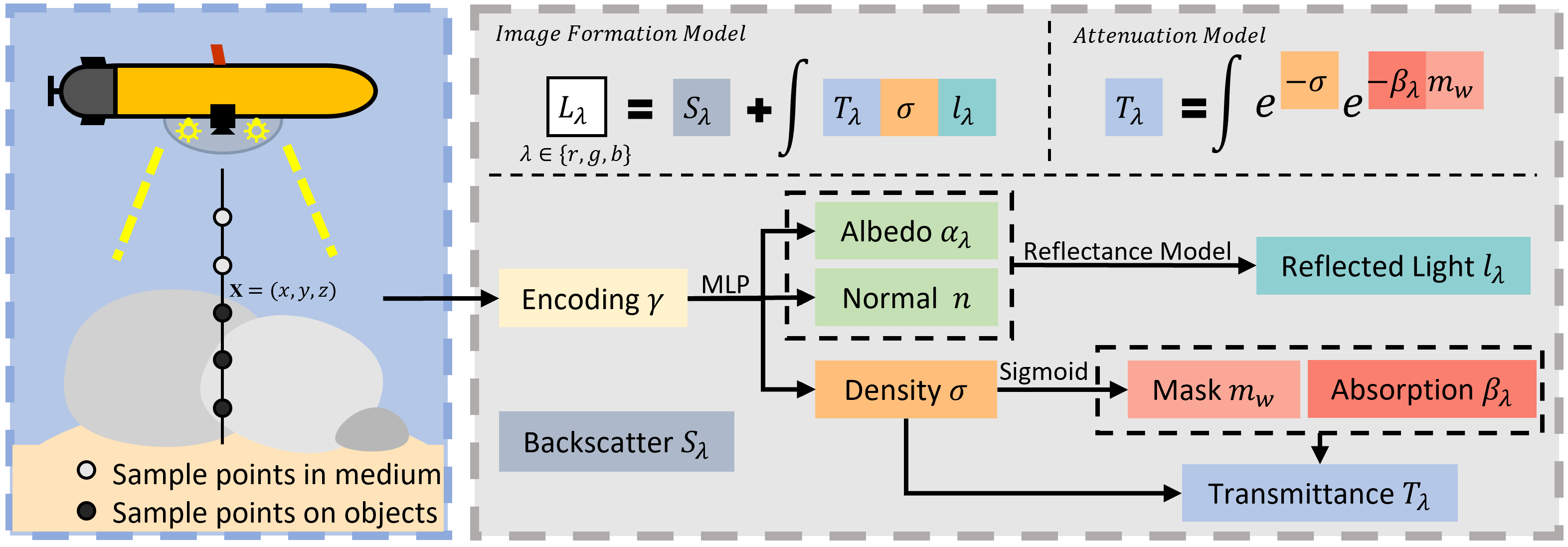}
    \end{subfigure}%
    \caption{Our proposed model: Sample points $\mathbf{x}$ are first mapped into positional encoding $\gamma(\mathbf{x})$, as the input of an MLP. The output of the MLP consists of albedo $\boldsymbol{\alpha}$, surface normal $\mathbf{n}$, and volume density $\sigma$. Backscatter $S_\lambda$ and attenuation coefficient $\beta_\lambda$ are global parameters optimized along with the MLP. With $\boldsymbol{\alpha}$ and $\mathbf{n}$ we can calculate the reflected radiance $l_\lambda$ from the scene. We apply a sigmoid function on $\sigma$ to separate water from scene and calculate transmittance $T_\lambda$ through the scene and water using different coefficients. With $S_\lambda$, $T_\lambda$, $\sigma$ and $l_\lambda$, our rendering model predicts the pixel values in the image.} \label{method:pipeline}
\end{figure*}
\par We employ neural reflectance field~\cite{neuralref} to model the underwater scene observed by an underwater robot with onboard lights.
The continuous scene is represented as a function of 3D location $\mathbf{x} = (x,y,z)$  in the global coordinate frame. The outputs of the function are the rendering properties $(\sigma, \boldsymbol{\alpha}, \mathbf{n})$, where $\sigma$ is the volume density, $\boldsymbol{\alpha}=(\alpha_r, \alpha_g, \alpha_b)$ is the albedo and $\textbf{n} = (n_x, n_y, n_z)$ is the surface normal (see Fig.~\ref{method:pipeline}).

\par In practice, we first sample 3D points $\mathbf{x}$ on camera rays in the global coordinate frame.
We then use hash encoding $\gamma$ to map the input $\mathbf{x}$ into a higher-dimensional space~\cite{mueller2022instant} before feeding it into a nested \ac{MLP}:
\begin{equation}
    (\sigma, \boldsymbol{\alpha}, \mathbf{n}) = \mathrm{MLP}(\gamma(\mathbf{x}))
\end{equation}

\subsection{Rendering Equations}

The volume rendering equation~\cite{pharr2016PhysBR, fong2017production} maps a camera ray $\mathbf{x} = \mathbf{o}-t\boldsymbol{\omega}$ into {the radiance $L_\lambda$ captured at location $\mathbf{o}$ in direction $\boldsymbol{\omega}$:}
\begin{equation}
    \begin{aligned}
        L_\lambda(\mathbf{o}, \boldsymbol{\omega})=\int_{t=0}^d T_\lambda(\mathbf{x}) \sigma(\mathbf{x}) {l}_\lambda(\mathbf{x}) dt\\
    \end{aligned}
    \label{eq:render}
\end{equation}
Here $T_\lambda$ is the transmittance from $\mathbf{x}$ to $\mathbf{o}$, $\sigma$ is the volume density, $l_\lambda$ is the scattered radiance from $\mathbf{x}$ to $\mathbf{o}$ along the ray, and $\lambda$ indicates the wavelength. In this study, the wavelength is discretized into RGB space that $\lambda\in \{r,g,b\}$~\cite{akkaySpace}.

For a light beam emitted from $\mathbf{x}$ to $\mathbf{o}$, the fraction of light that reaches the camera is described by the transmittance $T_\lambda$:
\begin{equation}
            T_\lambda(\mathbf{x})= \textrm{exp}({-\int_{s = 0}^t {\sigma}_\lambda(\mathbf{o}-s\boldsymbol{\omega})ds})
            \label{eq:transmittance}
\end{equation}
Here, $\sigma_\lambda$ denotes the attenuation coefficient as a function of the 3D location $\mathbf{o}-s\boldsymbol{\omega}$, which combines the extinction of light due to both volume-density-dependent out-scattering and wavelength-dependent absorption~\cite{jaffe1990, fong2017production}. The formulation of $\sigma_\lambda$ will be further discussed in~\ref{sec:watereff}.

\par The scattered radiance $l_\lambda$ from the scene, as a part of the integrand in Eq.~\ref{eq:render}, is formulated as follows:
\begin{equation}
\begin{aligned}
    l_\lambda(\mathbf{x}) &= \int_{S^2} f_\lambda(\mathbf{x}, \boldsymbol{\omega}, \boldsymbol{\omega}_i)I_\lambda(\mathbf{x}, \boldsymbol{\omega}_i) d\boldsymbol{\omega}_i
\end{aligned}
\label{eq:scatterredrad}
\end{equation}
where $S^2$ represents the spherical domain around point $\mathbf{x}$, $f_\lambda$ is the phase function that governs the distribution of light scattered at $\mathbf{x}$, and $I_\lambda$ is the incident radiance from direction $\boldsymbol{\omega}_i$ into $\mathbf{x}$.
\par In practice, we follow the assumptions in~\cite{ccbryson} that object surfaces underwater are Lambertian, which {scatter} light {in} all directions equally. Following Lambert's cosine law, the phase function for objects underwater is described as: $f_\lambda(\mathbf{x},\boldsymbol{w},\boldsymbol{w}_i) = \alpha_\lambda(\mathbf{x})\cos(\mathbf{n}(\mathbf{x}),\boldsymbol{\omega}_i)$. Here ${\alpha_\lambda}(\mathbf{x})$ and $\mathbf{n}(\mathbf{x})$ are the albedo and normal at $\mathbf{x}$ estimated by the neural network. In other words, we are not modeling any specular reflection which is rare underwater.

\par Inferring the the phase function $f_\lambda$ of water volumes, i.e. \ac{VSF}, is challenging and not scalable on real robots due to different light and camera configurations. To address this, we propose approximating backscatter in the image as a constant and moving away from estimating \ac{VSF} (see~\ref{subsec:approx}), by which the complexity of our approach is significantly reduced while still achieving accurate and realistic rendering results.

\par Similar to~\cite{ccbryson}, we only consider direct illumination from onboard lights. While natural and ambient light also impacts the lighting in shallow water, they are out of the scope of this work. The direct illumination on point $\mathbf{x}$ from the light source is expressed by: 
\begin{equation}
\begin{aligned}
    I_\lambda(\mathbf{x}, \boldsymbol{\omega}_i) &= T^i_\lambda(\mathbf{x})E^i_\lambda(\mathbf{x})
\end{aligned}
\end{equation}
 Here {$i$ indicates the light source from direction $\boldsymbol{\omega}_i$, $T^i_\lambda$ is the transmittance from the light source to $\mathbf{x}$} (the calculation is similar to Eq.~\ref{eq:transmittance}), and {$E^i_\lambda(\mathbf{x})$ is the intensity of light source $i$ evaluated at $\mathbf{x}$} taking light fall-off with distance into account.

\subsection{Unified Transmittance Model}
\label{sec:watereff}
\par The attenuation of light in water can be modeled with a {transmittance term $T_\lambda$ given attenuation coefficient $\sigma_\lambda$ and distance $t$}:
\begin{equation}
    T_\lambda = \textrm{exp}({-\int_{s = 0}^t {\sigma}_\lambda ds})=\textrm{exp}(-\sigma_\lambda t)
    \label{eq:Tsim}
\end{equation}
Given the emitted radiance $E$, the arrived radiance is $T_\lambda E$. The attenuation coefficient $\sigma_\lambda$ for water can be decomposed into the {out-scattering coefficient $\sigma$ and the absorption coefficient $\beta_\lambda$}~\cite{jaffe1990}. Notably, the out-scattering coefficient $\sigma$ is independent of the wavelength of the light~\cite{akkayFormation}, and can be represented as the volume density in rendering equations.
\par In the neural reflectance field, volume density is a function of spatial location $\mathbf{x}$, so we have:
\begin{equation}
    \sigma_\lambda(\mathbf{x}) = \sigma(\mathbf{x})+\beta_\lambda
    \label{eq:siglambdasim}
\end{equation}
where $\sigma(\mathbf{x})$ is predicted by the neural implicit functions and $\beta_\lambda$ will be optimized as a global parameter that doesn't change with spatial locations.
\par On a camera ray, points in the water attenuate light through both absorption and out-scattering, as described by Eq.~\ref{eq:siglambdasim}. On the contrary, points on objects have no wavelength-dependent absorption effects. So for underwater scenes $\sigma_\lambda(\mathbf{x})$ can be formulated as follows:
\begin{equation}
    \sigma_\lambda(\mathbf{x}) = \begin{cases}
\sigma(\mathbf{x})+\beta_\lambda,& \text{if $\mathbf{x}$ is in water} \\
\sigma(\mathbf{x}), & \text{if $\mathbf{x}$ is on objects}
\end{cases}
\label{eq:switchcase}
\end{equation}
When sampling points from non-transparent objects, the volume density $\sigma(\mathbf{x})$ should typically be large enough that regardless of whether $\mathbf{x}$ is in water or on objects, $\sigma(\mathbf{x})\approx \sigma(\mathbf{x})+\beta_\lambda$. However, it is still important to maintain the separate attenuation coefficients in Eq.~\ref{eq:switchcase} during training until the prediction of $\sigma(\mathbf{x})$ has converged.
\par To apply Eq.~\ref{eq:switchcase}, we need to differentiate water from the rest of the scene. We experimentally observe that the value of $\sigma(\mathbf{x})$ for objects is at least 10 times greater than that in clear water.
 This observation also aligns with the measurements by Jerlov~\cite{jerlov1976marine}.
Assuming that there are no highly transparent objects in the scene other than water, we define the following logistic regression functions using the sigmoid function:
\begin{equation}
    \begin{aligned}
        m_o(\mathbf{x})&=\textrm{sigmoid}(a(\sigma(\mathbf{x})-b))\\
        m_w(\mathbf{x})&=1-m_o(\mathbf{x})
    \end{aligned}
    \label{eq:sigmoid1}
\end{equation}
where {$m_o$ and $m_w$ indicate the probabilities of the query point $\mathbf{x}$ being on non-transparent objects and water}, respectively. Specifically, $a$ controls the steepness of the sigmoid function, and a higher value of $a$ results in higher confidence in prediction, but it may also increase the risk of vanishing gradient. $b$ determines the density threshold used to distinguish water from objects. With $m_o$ and $m_w$, we can express $\sigma_\lambda(\mathbf{x})$ in the following form:
\begin{equation}
\begin{aligned}
        \sigma_\lambda(\mathbf{x}) &= m_w(\mathbf{x})(\sigma(\mathbf{x})+\beta_\lambda)+m_o(\mathbf{x})\sigma(\mathbf{x})\\ &= \sigma(\mathbf{x})+m_w(\mathbf{x})\beta_\lambda
\end{aligned}
\label{eq:siglambda}
\end{equation}
In other words, $m_o$ and $m_w$ can be considered as masks on sample points, exposing those in the water and objects to distinct light-transmitting physics. 

\subsection{Approximating Water Effects}
\label{subsec:approx}
\begin{figure}[t]%
    \centering
    \begin{subfigure}{.45\textwidth}
    \includegraphics[width=\linewidth]{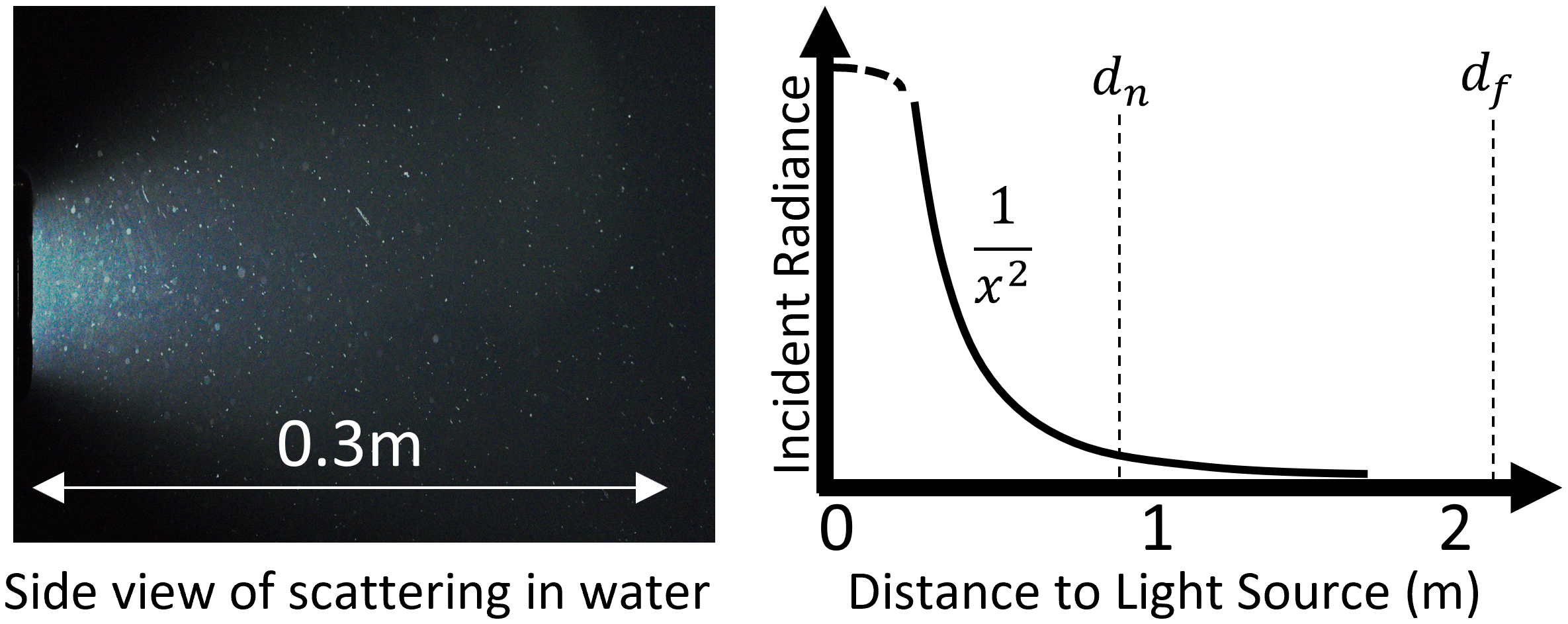}
    \end{subfigure}%
    \caption{A side view of scattering generated from an LED light (left) reflects the intensity distribution of incident radiance. We observe significant light fall-off with the distance from the light source. The plot on the right sketches a typical light fall-off curve. $d_n$ and $d_f$ indicate the typical positions of near and far bounding planes. When the distance is close to the dimensions of the lighting component, we need to precisely calibrate the lighting and imaging components to approximate the curve. The rest of the curve can be approximated with the inverse-square law.}%
    \label{fig:scatter}
\end{figure}

The backscatter effects in water can be described using a \ac{VSF}.
However, in learning neural scene representations from real underwater data, we encounter difficulties in modeling \acp{VSF}. Firstly, backscatter from the closer regions of the field of view has a greater impact {on} imaging (Fig.~\ref{fig:scatter}). We need a precise imaging system model to accurately infer the \ac{VSF} in this area. This requires detailed information about the dimensions and poses of the camera and light source. However, calibrating such a system complicates the deployment of our algorithm on real robots and is hard to scale across different robot platforms. Secondly, estimating the \ac{VSF} along the ray prevents us from using bounding planes, which could significantly enhance the sampling efficiency and avoid overfitting by constraining the viewing frustum from multiple views.
To address the issues mentioned above, we propose several approximations to avoid modeling \acp{VSF}:
    \subsubsection{Backscatter as a constant~\label{app:app1}} The backscatter captured in the image can be approximated as a constant $S_\lambda$, as the majority of backscatter comes from the region close to the light source, which is not affected when the images are taken from different depths and perspectives (see Fig.~\ref{fig:scatter}).
    \subsubsection{Co-centered camera and light source\label{app:app2}} Points are only sampled between the near and far bounding planes, and their distances to the camera are sufficiently large compared to the typical dimensions of the components of the imaging system. Therefore, we model the light source as a single point light source that is co-centered with the camera, similar to \cite{neuralref}. We use the inverse-square law to calculate the incident radiance $E_\lambda(\mathbf{x})$.


\par We design a loss function that enforces the model to output $\sigma(\mathbf{x}) = 0$ if $\mathbf{x}$ is in water (see~\ref{subsec:loss}). With this constraint, we can avoid double-count backscatter with both $S_\lambda$ and Eq.~\ref{eq:render} since the integrand in Eq.~\ref{eq:render} will have zero values for $\mathbf{x}$ in the water. Additionally, constraining $\sigma(\mathbf{x}) = 0$ for $\mathbf{x}$ in water allows us to calculate the attenuation between the near bounding plane and the camera without sampling points. As a parameter to be optimized in training, $\beta_\lambda$ will approach $\sigma_\lambda(\mathbf{x})$ when $\sigma(\mathbf{x})$ approaches 0 according to Eq.~\ref{eq:siglambdasim}. Then the transmittance between the near bounding plane and the camera will be $T_\lambda^n = \textrm{exp}(-\beta_\lambda d_n)$ according to Eq.~\ref{eq:Tsim}, and Eq.~\ref{eq:render} can be written as:
\begin{equation}
    \begin{aligned}
        L_\lambda(\mathbf{o}, \boldsymbol{\omega})=S_\lambda+T_\lambda^n\int_{t=d_n}^{d_f} T_\lambda(\mathbf{x}) \sigma(\mathbf{x}) {l}_\lambda(\mathbf{x}) dt\\
    \end{aligned}
    \label{eq:render2}
\end{equation}
Here $d_n$ and $d_f$ are the distances from the camera to near and far bounding planes respectively.

\subsection{Ray Marching}
\label{subsec:raymarch}
We numerically estimate Eq.~\ref{eq:render2} by ray marching. Rays are sampled from the center of the camera and pass through uniformly sampled points on the image plane in training. Points are then sampled along the ray between the near and far bounding planes. The rendering equation is discretized as follows:
\begin{equation}
    \begin{aligned}
        L_\lambda(\mathbf{o}, \boldsymbol{\omega}) &= S_\lambda + T_\lambda^n\sum\nolimits_{i = 0}^N T_\lambda(x_i)\Phi_\lambda(x_i) l_\lambda(x_i)\\
        T_\lambda(x_i) &= \textrm{exp}(-\sum\nolimits_{j = 0}^i \sigma_\lambda(x_j)\delta_j) \\
        \Phi_\lambda(x_i)&=\frac{\sigma(x_i)}{\sigma_\lambda(x_i)}(1-\textrm{exp}(-\sigma_\lambda(x_i)\delta_i))
        \\
        l_\lambda(x_i) &= T_\lambda^n T_\lambda(x_i) E_\lambda(x_i)\alpha_\lambda\cos(\textbf{n}(x_i),\boldsymbol{\omega})
    \end{aligned}  
    \label{eq:raymarch1}
\end{equation}
where $\delta_i$ denotes the step size at sample point $x_i$. It is worth noticing that transmittance terms $T_\lambda^n$ and $T_\lambda(x_i)$ are used in both the calculation {of} incident radiance $l_\lambda$ and the sensed radiance $L_\lambda$ according to approximation~\ref{app:app2}.
\par The opacity $\Phi_\lambda$ corresponds to the term $1-\textrm{exp}(\sigma(x_i)\delta_i)$ in NeRF and its variants. In NeRF, the volume density $\sigma$ governs both the emission and attenuation of the radiance, making it sufficient to model objects in the air, haze, and even transparent glowing gas~\cite{renderingMax1995}.
In our study, we need to model the wavelength-dependent attenuation, which requires both the volume density $\sigma$ and the attenuation coefficient $\sigma_\lambda$ to play a role together in $\Phi_\lambda$. However, if $\sigma_\lambda$ in the denominator approaches $0$ in training, the model will encounter numerical issues. To avoid this, we take advantage of our proposition in~\ref{subsec:approx} that enforces $\sigma(x_i)=0$ if $x_i$ is in the water, so $\Phi_\lambda(x_i) = 0 = 1-\textrm{exp}(-\sigma(x_i)\delta_i)$.
When $x_i$ falls on objects, $\sigma_\lambda(x_i)=\sigma(x_i)$ according to Eq.~\ref{eq:siglambda}, so $\Phi_\lambda(x_i)=1-\textrm{exp}(-\sigma(x_i)\delta_i)$. We then simplify $\Phi_\lambda(x_i)$ into the following form, which is identical to the opacity term in NeRF~\cite{mildenhall2020nerf}:
\begin{equation}
    \Phi_\lambda(x_i)=1-\textrm{exp}(-\sigma(x_i)\delta_i)
\end{equation}


\subsection{Loss Function}
\label{subsec:loss}
We use $L_2$ loss to optimize the rendered radiance with captured pixel values from the raw image, which has linear color. As a result, the $L_2$ loss will be dominated by errors in the brighter parts of the image, and the darker parts will have low rendering quality. To achieve better visual results, we apply a stronger penalization on errors in the darker parts of the image by tone-mapping $\psi$ on both the model output and raw pixel values before passing them into the loss function {$\mathcal{L}$} as suggested by~\cite{mildenhall2022rawnerf}:
 \begin{equation}
    \mathcal{L} = \sum_\lambda\sum_{r\in R} \lVert \psi(\hat{L}_\lambda(r)) -\psi(L_\lambda(r))\rVert_2^2
     \label{eq:lossfunc}
 \end{equation}
 Here $R$ is the sampled ray batch, $\hat L$ is the raw pixel value and $L$ is the radiance predicted from the model. We use the gamma correction proposed in~\cite{adobesrgb} as our $\psi$ function to map the linear color to sRGB space.
\par As proposed in~\ref{subsec:approx}, we want to constrain the volume density $\sigma(\mathbf{x})=0$ for $\mathbf{x}$ {in water}. We first set $\sigma(\mathbf{x})=0$ for $\mathbf{x}$ in water by multiplying $m_o(\mathbf{x})$.
This gives us the refined volume density $\Bar{\sigma}(\mathbf{x})$:
\begin{equation}
    \Bar{\sigma}(\mathbf{x}) = m_o(\mathbf{x})\sigma(\mathbf{x})\end{equation}
Then we are able to calculate the refined radiance $\Bar{L}_\lambda(r)$ with equations in~\ref{subsec:raymarch} using $\Bar{\sigma}(\mathbf{x})$ in the place of ${\sigma}(\mathbf{x})$. The refined loss {$\Bar{\mathcal{L}}$} is calculated similarly to Eq.~\ref{eq:lossfunc}:
\begin{equation}
    \Bar{\mathcal{L}} = \sum_\lambda\sum_{r\in R} \lVert \psi(\hat{L}_\lambda(r)) -\psi(\Bar{L}_\lambda(r))\rVert_2^2
\end{equation}

The total loss is $\mathcal{L}_{total} = \mathcal{L} + \Bar{\mathcal{L}}$. By optimizing $\mathcal{L}_{total}$, we are encouraging the model to generate the same results with $\sigma(\mathbf{x})$ and $\Bar{\sigma}(\mathbf{x})$. So the prediction of $\sigma(\mathbf{x})$ from network will converge to $\Bar{\sigma}(\mathbf{x})$, where for $\mathbf{x}$ in the water, $\sigma(\mathbf{x})=0$.

\subsection{Re-rendering with True Color}
To re-render the image with true color, we just need to remove the backscatter $S_\lambda$,  wavelength-dependent absorption $\beta_\lambda$ and volume density $\sigma(x)$ for $\mathbf{x}$ in water. We only need to use $\Bar{\sigma}(\mathbf{x})$ in calculating transmittance $T$ and opacity $\Phi$. The rendering equation in~\ref{subsec:raymarch} becomes the following:
\begin{equation}
    \begin{aligned}
        L_\lambda(\mathbf{o}, \boldsymbol{\omega}) &= \sum\nolimits_{i = 0}^N T(x_i)\Phi(x_i) l_\lambda(x_i)\\
        T(x_i) &= \textrm{exp}(-\sum\nolimits_{j = 0}^i \Bar{\sigma}(x_j)\delta_j) \\
        \Phi(x_i)&=1-\textrm{exp}(\Bar{\sigma}(x_i)\delta_i)
        \\
        l_\lambda(x_i) &=  T(x_i) E_\lambda(x_i)\alpha_\lambda\cos(\mathbf{n}(x_i),\boldsymbol{\omega})
    \end{aligned}  
    \label{eq:raymarch2}
\end{equation}

\section{EXPERIMENTS}
\label{sec:experiments}
\subsection{Dataset}

\begin{figure}[t]%
    \centering
    \begin{subfigure}{.48\textwidth}
    \includegraphics[width=\linewidth]{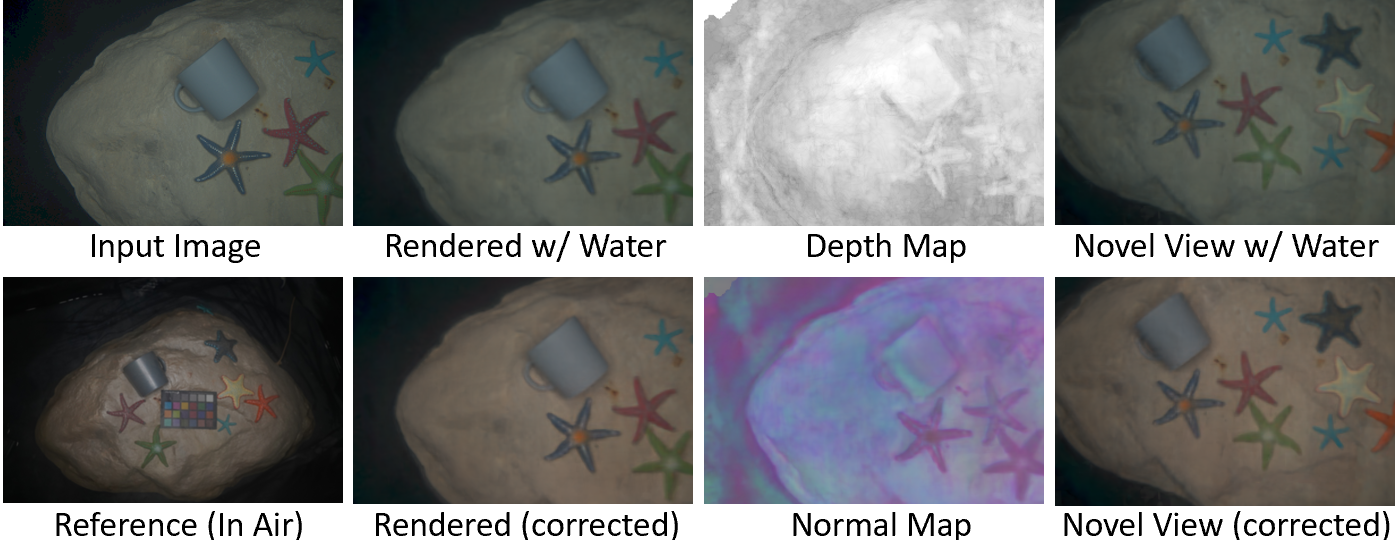}
    \end{subfigure}%
    \caption{An example of our color correction algorithm being applied on the data collected in a water tank. Visualization show that our proposed method is able to recover the color and geometry of the underwater scene, and render image with consistant appearance from novel views.}
    \label{fig:viz}
\end{figure}

\begin{figure*}[t]%
\centering
    \begin{subfigure}{.95\textwidth}
    \includegraphics[width=\linewidth]{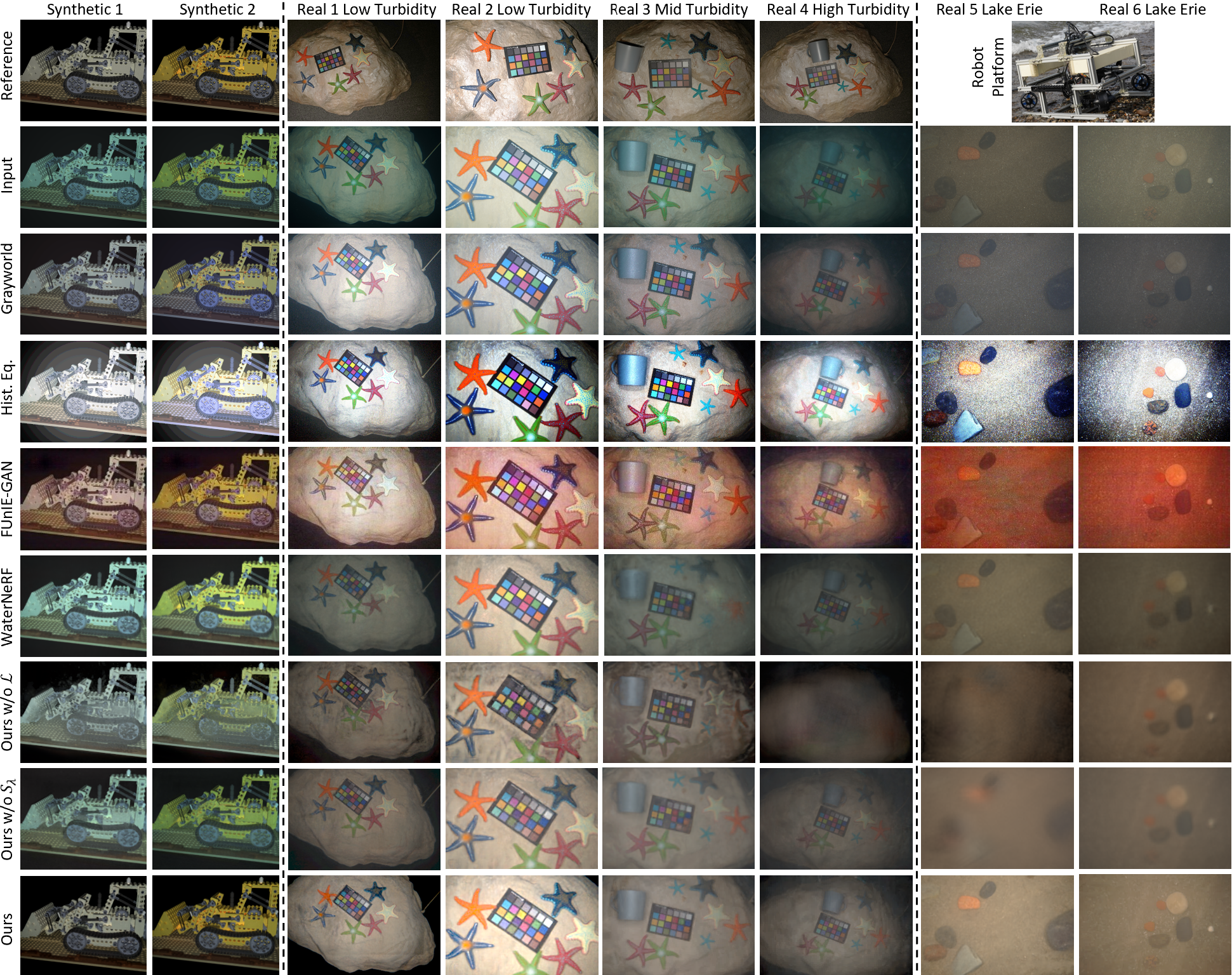}
    \end{subfigure}%
     \caption{Visualizations of color restoration. For good visualization quality, real images are visualized in sRGB space. {Reference images for synthetic data are generated by rendering without any water effects. Reference images for Real 1-4 (water tank) data are captured with same camera and light setup in the air.}}%
    \label{exp:main}
\end{figure*}

We collect our underwater data in a water tank with 1.3m water depth (An example is shown in Fig.~\ref{fig:viz}), {as well as in natural water (Lake Erie) with our underwater robot platform}.
Our imaging system consists of a Sony ILCE-7M3 camera with a 40mm prime lens and LED lights.
The imaging system is housed in a waterproof case and fully submerged when collecting data. The images are captured using 1/250s exposure time, $f/5.6$ aperture, and ISO 1600. The raw image files with 14-bit pixel values in HDR space are decoded, denoised, and scaled into 8-bit images with linear values using RawPy~\cite{rawpy}.
We placed artificial decorations with various colors on the bottom of the water tank together with a Macbeth ColorChecker~\cite{McCamy-1976-CC}. We use the manufacturer's (X-rite) software
to balance the image color as ground truth {in quantitative study}, which is only used for comparison purposes and does not play a role in our proposed algorithm. We acquire camera poses from COLMAP~\cite{colmap} with post-processed JPEG images to ensure high feature quality.
We also build our synthetic data based on implementations from~\cite{mcraytracergithub, uwroborender} and measurements from ~\cite{jerlov1976marine, petzold1972volume}.



\subsection{Implementations}
\par Our code is developed using PyTorch3D Library~\cite{ravi2020pytorch3d}.
We use hash encoding proposed in iNGP~\cite{mueller2022instant} for positional encoding. We choose $a = 3$ and $b=3$ empirically for our sigmoid function in Eq.~\ref{eq:sigmoid1} {(same $a$, $b$ values are used for all experiments)}. Our neural implicit function consists 3 sub-\acp{MLP} predicting $\sigma$, $\boldsymbol{\alpha}$, and $\mathbf{n}$ respectively similar to $S^3$-NeRF~\cite{yang2022s3nerf}. We use \texttt{LeakyReLU} between consecutive linear layers and \texttt{SoftPlus} as the final layer in predicting $\sigma$ and $\boldsymbol{\alpha}$ to guarantee non-negative outputs.
\par The model is trained on an Nvidia RTX 4090 GPU with 24GB memory. In each training iteration, we sample 1000 rays from one image and 100 points on each ray. The model is trained for $50k$ epochs for each scene. 

\subsection{Comparisons}
We compare our results with grayworld algorithm~\cite{grayworld}, histogram equalization~\cite{histeq}, FUnIE-GAN~\cite{funniegan} and WaterNeRF~\cite{Sethuraman2022WaterNeRFNR} (we use open-sourced Sinkhorn loss implementation in GeomLoss Library~\cite{geomloss}). Results are shown in Fig.~\ref{exp:main}. Grayworld algorithm and histogram equalization algorithm only correct color well on Synthetic 1 sequence, in which the object's albedo is dominated by low-saturation colors. Under such circumstances, the grayworld and histogram-equalizing assumptions align well with the true color distribution of the scene, so they perform well. However, when we change the body color of the bulldozer to bright yellow (Synthetic 2), grayworld algorithm and histogram equalization are getting downgraded as their assumptions fail. We can observe the same in real images where the albedo of the scene is dominated by a sand-colored background. Grayworld and histogram equalization algorithms both tend to balance it into grey color. We also observe that both predictions from grayworld and histogram equalization algorithm unpredictably add more veiling light effects or noises into the raw image.

\begin{table}[t]
\caption{MSE {of A/B channels} in CIELAB Space $\downarrow$ (pixel values range 0-255)}
\centering
\begin{tabular}{c@{\ }c@{\ }c@{\ }c@{\ }c@{\ }c@{\ }c}
\hline
\textbf{} & Syn. 1 & Syn. 2 & {Real 1} & {Real 2} & {Real 3} & {Real 4} \\ 
\cline{2-7}
  & A / B  & A / B  & A / B & A / B & A / B  & A / B \\ \hline
Grayworld &  4.66/19.7 & 22.1/91.4 & 10.7/82.3 & \textbf{15.1}/110 &{18.5/104}&{5.59/9.06} \\
Hist.\ Eq.& 5.94/23.7 & 21.5/63.5 & 15.7/87.0 & 43.6/110 & {23.6/102}&{18.8/31.7}\\
FUnIE-GAN & 75.5/30.0  & 61.2/36.1 & 33.8/61.5 & 97.4/49.5& {95.2/\textbf{40.3}}& {103/61.7}\\
WaterNeRF & 55.7/10.3  & 60.2/13.7  & 20.9/77.6 & 15.4/31.1& {39.5/96.3}&{10.1/3.73}\\
{Ours w/o $\bar{\mathcal{L}}$} & {22.8/18.8} & {60.8/41.1} & {14.1/74.0} & {21.1/47.7} & {21.0/91.4} & {4.55/4.53}\\
{Ours w/o $S_\lambda$} &{45.3/8.73}&{71.8/30.3}&{11.6/70.0}&{20.6/41.3}&{19.3/92.4} & {\textbf{2.82}/\textbf{3.06}}
\\Ours & \textbf{1.15}/\textbf{2.39} &  \textbf{4.08}/\textbf{9.01} &\textbf{9.68}/\textbf{42.5} & 19.2/\textbf{30.4} & {\textbf{18.2}/89.9} & {3.21/3.96}\\ \hline
\end{tabular}
\label{tab:psnr}
\end{table}

\begin{table}[t]
\centering
\caption{Angular Error in sRGB Space $\downarrow$ (radians)}
\begin{tabular}{c@{\ \ \ }c@{\ \ \ }c@{\ \ \ }c@{\ \ \ }c@{\ \ \ }c@{\ \ \ }c}
\hline
& Syn. 1 & Syn. 2 & Real 1 & Real 2 & Real 3 & Real 4 \\ \hline
Grayworld & 0.0724 & 0.2186 & 0.1381 & 0.0962 & {0.1299} & {0.0563}\\
Hist. Eq. & 0.0758 & 0.2482 & 0.1421 & 0.1916 & {0.1569} & {0.0810} \\
FUnIE-GAN & 0.1107 & 0.1166 & 0.1221 & 0.1655 & {0.1607} & {0.1680}\\
WaterNeRF & 0.1403 & 0.1748 & 0.1303 & 0.0596 & {0.1537} & {0.0514} \\
{Ours w/o $\bar{\mathcal{L}}$} & {0.1048} & {0.2215} & {0.0938}&{0.0759}&{0.1246} & {0.0803}\\
{Ours w/o $S_\lambda$} &{0.1121}&{0.2099}&{0.0909}&{0.0728}&{0.1243} & {\textbf{0.0342}} \\
Ours & \textbf{0.0361} & \textbf{0.0458} & \textbf{0.0837} & \textbf{0.0591} & {\textbf{0.1119}} & {0.0404}\\ \hline
\end{tabular}
\label{tab:angularerror}
\end{table}

\par As one of the latest GAN-based methods, FUnIE-GAN is pre-trained on annotated underwater images. In our experiments, we find FUnIE-GAN overshooting in the red channel as shown in Fig.~\ref{exp:main}, implying that color distributions in their training data are less red than ours. In other words, instead of naive assumptions such as histogram equalization, GAN-based methods learn a color distribution from pre-collected datasets. The inherent color distribution in the data for pretraining can deviate from observations as well.
\par WaterNeRF tackles the problem by applying the physical constraints from Jaffe-McGlamery model while approaching the histogram-equalized image. We acknowledge that it's not a fair comparison since WaterNeRF works for any kind of illumination while our data is only for situations where light and camera move as a rigid body. We observe that when the histogram-equalized image is flawed, e.g. with our synthetic data, WaterNeRF can be significantly downgraded. We also find our method outperforms WaterNeRF in color consistency on real data. For example, comparing Real 1 and 2 images in Fig.~\ref{exp:main}, which is from the same image sequence, our method restores the color of the rock with better consistency since we model the albedo and light reflection of the scene, while WaterNeRF only models constant radiance, which fails when the light source moves. In general, from the comparisons, our method restores color in both synthetic and real-world data with the most consistent performance.


\par We present two metrics for quantitative evaluation: \ac{MSE} of {A/B channels in CIELAB space} (Table~\ref{tab:psnr}) and mean angular error~\cite{Sethuraman2022WaterNeRFNR} in the sRGB space (Table~\ref{tab:angularerror}). CIELAB is designed to approximate human vision in a uniform space~\cite{ISO15076-1:2010} and sRGB is the standard colorspace in which the image is presented.
For the synthetic dataset, we use the ground truth from the renderer and calculate both metrics directly. For {water tank} data, {we use color checker software corrected images as ground truth.}
\par As revealed by \ac{MSE} (Table~\ref{tab:psnr}), our method performs the best on synthetic data on {both A/B} channels.
{Among} the 4 real images evaluated, {our method overall performs well on scenes with low to medium turbidity (Real 1-3). With high water turbidity (Real 4), our method is downgraded as our approximation of backscatter is not sufficient to handle such water condition.}
Besides evaluating LAB channels separately, angular error (Table~\ref{tab:angularerror}) reflects the color similarity in the entire RGB space. Results show that our method performs best across all data {except the high-turbidity data}, which is consistent with the visualizations in Fig.~\ref{exp:main}.


\par {We did ablation study on our proposed refined loss $\mathcal{\bar{L}}$ and backscatter term $S_\lambda$.
Results are included in Fig.~\mbox{\ref{exp:main}}, Table~\mbox{\ref{tab:psnr}} and Table~\mbox{\ref{tab:angularerror}}.
We found that without $\mathcal{\bar{L}}$, the quality of image reconstruction is severely reduced when water effects are removed. This implies that light scattered from the scene and from water is mixed up in the model learned, so when water effects are being removed from the reconstruction, the image appears to be incomplete. We also observe that without $S_\lambda$, quantitative results show downgraded model performance in most testing cases. The only exception is the high-turbidity case (Real 4) in which our approximation of backscatter is not sufficient to properly model heavy scattering effects, thus underperforming the model without $S_\lambda$.}

\section{DISCUSSIONS}
\label{sec:discussions}


\par This work is directly applicable to underwater imagery collected when dominant light sources move with the camera as a rigid body, such as in deep water, ice-covered water, or cave water. However, it may fail in the following scenarios:
\begin{itemize}
	\item When the light source is a combination of onboard strobes (point light sources), natural light and ambient light, our model is inadequate for accurately representing water effects from mixed light sources.
	\item In the presence of highly turbid and layered water, scattering effects vary more significantly with depth, and the robot will have to observe the scene at a closer range (breaking approximation~\ref{app:app1}). Modeling backscatter as a constant could potentially lead to failure.
	\item When the baseline between the camera and onboard light source is long, creating shadows in the observed scene, our model, which assumes co-centered light and camera, cannot accurately represent shadows (breaking approximation~\ref{app:app2}). This issue also arises with robots equipped with multiple cameras or light sources.
\end{itemize}

\section{CONCLUSIONS}
\label{sec:conclusions}
\par This work proposes a unified framework that learns underwater neural scene representations together with water effects for underwater robotic imagery. We demonstrate that our method is capable of restoring the true color of the underwater scene with a sequence of observations from different ranges and perspectives. By approximating the backscatter and simplifying the ray tracing, we avoid estimating \ac{VSF}, which is numerically unstable and requires precise calibration of lighting and imaging system. Additionally, our proposed method generates dense results with end-to-end differentiability and does not rely on any pre-training, depth estimation, or assumptions on prior color distributions.
\par Future work will extend our model to address the issues discussed in~\ref{sec:discussions}. Our long-term goal is to achieve true color correction for all types of underwater lighting conditions.





\renewcommand{\bibfont}{\normalfont\small}
\printbibliography


\end{document}